\documentclass[twoside,leqno,twocolumn]{article}

\usepackage[letterpaper]{geometry}

\usepackage{ltexpprt}

\usepackage[square,sort,comma,numbers]{natbib}
\usepackage{caption} 
\usepackage{multirow}

\usepackage{array}
\newcolumntype{P}[1]{>{\centering\arraybackslash}p{#1}}

\usepackage{algorithm}
\usepackage{algorithmicx}
\usepackage[algo2e]{algorithm2e} 
\usepackage{enumitem}
\usepackage{subcaption}
\usepackage{amsmath}
\usepackage{amssymb}
\usepackage{xcolor}
\usepackage{graphicx}
\usepackage{grffile} 
\usepackage{footnote}
\usepackage{float}
\usepackage[switch]{lineno}

\usepackage{bbm}

\newtheorem{lem}{Lemma}
\newtheorem{thm}{Theorem}
\newtheorem{defn}{Definition}

\newtheorem{prop}{Proposition}

\begin{document}

\title{\Large Fair Representation Learning using Interpolation Enabled Disentanglement}
\author{Akshita Jha \thanks{Virginia Tech. akshitajha@vt.edu, reddy@cs.vt.edu}
\and Bhanukiran Vinzamuri \thanks{Amazon. ~vinzamub@amazon.com. Work done prior to joining Amazon.}
\and Chandan K. Reddy *}

\date{}

\maketitle


\fancyfoot[R]{\scriptsize{Copyright \textcopyright\ 2021\\
Unauthorized reproduction of this article is prohibited}}





\maketitle

\begin{abstract}
With the growing interest in machine learning community to solve real world problems, it has become crucial to uncover the hidden reasoning behind their decisions by focusing on the fairness and auditing the predictions made by these black-box models. In this paper, we propose a novel method to address two key issues: (a) Can we simultaneously learn fair disentangled representations while ensuring the utility of the learned representation for downstream tasks, and (b) Can we provide theoretical insights into when the proposed approach will be both fair and accurate. To address the former, we propose the method \textbf{FRIED}, \textbf{F}air \textbf{R}epresentation learning using \textbf{I}nterpolation \textbf{E}nabled \textbf{D}isentanglement. In our architecture, by imposing a critic-based adversarial framework, we enforce the interpolated points in the latent space to be more realistic. This helps in capturing the data manifold effectively and enhances utility of the learned representation for downstream prediction tasks. We address the latter question by developing theory on fairness-accuracy trade-offs using classifier-based conditional mutual information estimation. We demonstrate the effectiveness of FRIED on datasets of different modalities --- tabular, text, and image datasets. We observe that the representations learned by FRIED are overall fairer in comparison to existing baselines and also accurate for downstream prediction tasks. Additionally, we evaluate FRIED on a real-world healthcare claims dataset where we conduct an expert aided model auditing study providing useful insights into opioid addiction patterns.

\end{abstract}

\section{Introduction}\label{intro}
Fairness is an important component in building trustworthy AI systems for real-world problems. The other components of such a system compose of services for explainability and adversarial robustness. The goal of fair representation learning  problems~\cite{madras2018learning,zemel2013learning} has often been to formulate multi-objective optimization problems consisting of terms accounting for individual fairness, group fairness, utility, etc. In such a process, although we end up improving the fairness of the representation in accordance with the metric and the protected group we optimize for, we often incur a trade-off w.r.t. other performance metrics such as accuracy of the classifier. While ensuring fairness is critical in real-world problems, adversely affecting downstream classifier/clustering accuracy in the process is undesirable. In this regard, there is also a very limited understanding in the literature on what kinds of representations can mitigate this trade-off. Only recently emphasis has been laid on understanding and improving the trade-off using hypothesis testing, fairness-based empirical risk minimization, active and semi-supervised learning ~\cite{blum2019recovering,sanghamittrachernoff,noriega2019active,zhang2020fairness}.

In this paper, to address the problem of learning fair representations while also providing a theoretically sound mechanism of preserving the classifier accuracy, we propose FRIED, \textbf{F}air \textbf{R}epresentation learning using \textbf{I}nterpolation \textbf{E}nabled \textbf{D}isentanglement. We assume that we are explicitly provided with information about the protected group. The core method uses an autoencoder which tries to learn latent features which are disentangled from the protected group. 
This is accomplished using a critic network to disentangle the protected group information from the latent features. In addition, useful separability information is encoded within these latent features to aid with better downstream performance. Separability information here is learned while training the model by interpolating in the latent space to learn the data manifold effectively leveraging a separate interpolation critic. 
The detailed model description is provided in Figure~\ref{fig:architecture}.

An additional benefit of this approach is that the representation learned can be used for auditing black-box models to uncover the effect of the `proxy' features. A proxy feature is one that correlates closely with a protected attribute, and is causally influential in the overall behavior of the black-box model~\cite{yeom2018hunting}. This is of great interest to a policymaker who would like to understand the indirect influence of proxy features other than the protected attribute on the model outcome. To demonstrate this, in Section~\ref{sec:expertstudy}, we use FRIED and  audit a classifier used for predicting opioid addiction risk and validate the indirect influence insights obtained using an expert.

The rest of this paper is organized into the following sections. Section \ref{related} describes the related work. Section \ref{method} presents the required preliminaries needed to comprehend the proposed FRIED model and our methodology in detail. Section \ref{sec:friedtheory} elucidates how FRIED theoretically improves the fairness-accuracy trade-off. The experimental results are presented in Section \ref{exp} before concluding our work in Section \ref{concl}.

\section{Related Work}\label{related}
We divide the related work in this area into the following broad sections

\textit{Optimization-based}: LFR method was one of the first papers to propose an optimization-based formulation for fairness which encodes fidelity to the model prediction in addition to group fairness \cite{zemel2013learning}. A pre-processing-based model agnostic convex optimization framework to optimize for individual fairness, group fairness and utility was proposed by Calmon et al.~\cite{calmon2017optimized}.

\textit{Autoencoder-based}: Creager et al. \cite{creager2019flexibly} propose a method to learn a representation by disentangling the latent features w.r.t. multiple sensitive attributes. Sarhan et al. \cite{sarhan2020fairness} use entropy maximization to make the learned latent representation agnostic to the protected attributes. BetaVAE \cite{higgins2016beta} uses a modified VAE objective that encourages the reduction of the KL-divergenece to an isotropic Gaussian prior to ensure disentanglement of all latent codes. FactorVAE \cite{kim2018disentangling} encourages factorization of the aggregate posterior by approximating the KL-Divergence using the cross-entropy loss of a classifier. Conditional VAE \cite{sohn2015learning} improves the performance of VAE by conditioning the latent variable distribution over another variable. Louizos et al. \cite{louizos2015variational} proposed a variational fair autoencoder framework which investigates the problem of learning fair representations by treating protected attributes as nuisance variables that they purge from their the latent representations. 

\textit{Fairness-accuracy trade-off specific}:
Feldman et al. \cite{feldman2015certifying} demonstrate the fairness-utility trade-off by linking disparate impact to classifier accuracy. Zafar et al. \cite{zafar2017fairness} empirically demonstrate that their proposed methodology avoids disparate mistreatment at a small cost in terms of accuracy while Berk et al.~\cite{berk2017convex} experiment with varying weights of fairness regularizers to measure the `Price of Fairness'. Dutta et al. \cite{sanghamittrachernoff} use a mismatched hypothesis testing based framework to demonstrate that optimal fairness and accuracy can be achieved simultaneously. Wick et al. \cite{wick2019unlocking} show that fairness and accuracy can be in harmony by using semi-supervision and accounting for label and selection bias. Zhang et al. \cite{zhang2020fairness} use unlabeled data in a semi-supervised setting to improve the trade-off. Blum et al. \cite{blum2019recovering} propose the idea where fairness constraints can be combined with empirical risk minimization to learn the Bayes optimal classifier.

\textit{Model Auditing}: Xue et al. \cite{xue2020auditing} propose an auditing framework for individual fairness which allows an auditor to control for false discoveries. Marx et al. \cite{marx2019disentangling} propose the \emph{disentangling-influence} framework to learn a latent representation which is disentangled from the protected group and this representation is used for model auditing. There are similarities between FRIED and \emph{disentangling-influence}, however, their work is limited to model auditing alone, and it does not explore refining the latent representation by learning the manifold and improving the fairness-accuracy trade-off. Yeom et al. \cite{yeom2018hunting} study identifying proxy features in linear regression problems by solving a second order cone program.

\textit{Other Methods}: Early work by Kamiran et al. \cite{kamiran2012data} use pre-processing techniques to remove discrimination. Zhang et al. \cite{zhang2018mitigating} make use of adversarial training to maximize the predictor's ability to produce the correct output while simultaneously minimizing the adversary's ability to predict the protected attribute.

To the best of our knowledge, our proposed model is the the first one to employ interpolation to learn latent features, and empirically quantify using mutual information theory on how it mitigates the fairness-accuracy trade-off.

\section{Proposed Method}\label{method}

\subsection{Preliminaries}\label{background}

\begin{figure*}[h!]
  \includegraphics[width=\textwidth]{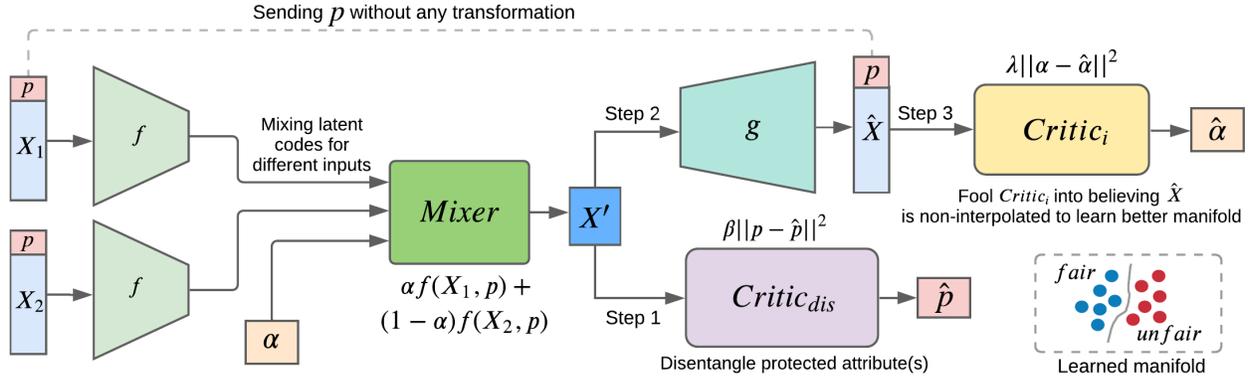}
  \caption{Architecture diagram, during training, of the proposed FRIED model that learns Fair Representation using Interpolation Enabled Disentanglement. The encoders $f$ share weights and take input pairs of instances $X_1$ and $X_2$, along with the protected attribute(s) $p$. The latent convex interpolation of the inputs from the $Mixer$ in Step 1 is input to the disentanglement critic $Critic_{dis}$, which predicts $\hat{p}$, and in Step 2, to the decoder $g$, along with $p$. In Step 3, the interpolation critic $Critic_i$ attempts to predict the interpolation coefficient $\hat{\alpha}$ from the reconstructed input $\hat{X}$. After the training is complete, the encoder $f$ can be used for inference and $g$ for reconstruction. Test time architecture is in the SM.}
  \label{fig:architecture}
\end{figure*}

Let $p$ denote the protected attribute(s), $X$ denote the input feature matrix, $Y$ denote the true label. Let $X_k$ denote the feature vector corresponding to instance $k$. $X'$ is the latent representation obtained after interpolation and $\hat{X}$ is the reconstructed input from the decoder. In order to learn fair disentangled representations, our system uses adversarially constrained interpolation along with training of autoencoders to efficiently disentangle factors of variations. In this section, we present a generalized view of the model. The complete train-time architecture of our model can be seen in Fig. \ref{fig:architecture}. The architecture for model auditing is shown in Fig.~\ref{fig:model_audit1}.

\subsection{\textbf{Learning Fair and Disentangled Representations}}\label{rep_learn}

The inputs, $X$ and $p$ are fed into the encoder $f$, which generates the latent representations for the input instances $(X_1, p)$ and $(X_2, p)$ by passing it through an encoding function, $f((X, p); \theta_f)$ where $\theta_f$ refers to the model parameters. After getting the latent representations, $f((X_1, p); \theta_f)$, and $f((X_2, p); \theta_f)$ are fed to the mixer. The mixer semantically mixes the latent codes as $X' = \alpha f(X_1, p) + (1-\alpha) f(X_2, p)$
where $\alpha \epsilon [0, 1]$. As a first step, this latent mixture is provided as an input to the adversary $Critic_{dis}$, responsible for disentangling the protected features. The adversary's sole purpose is to predict the protected attribute $\hat{p}$ from $X'$ without having any direct access to $p$. $Critic_{dis}$ ensures that once the training process converges, the latent code $X'$, does not contain any recoverable information about the sensitive attribute $p$. This helps in successfully disentangling $p$ from $X'$. In our formulation, we provide manifold information through latent interpolation for better disentanglement which is not explicitly provided in the formulation in Marx et. al. ~\cite{marx2019disentangling}. The loss function used to codify the disentanglement critic is as follows.

\begin{equation}
    L_{Critic_{dis}} = ||p - \hat{p}||_{2}^2
    \label{eq:critic_dis}
\end{equation}

where $\hat{p}$ is the output from $Critic_{dis}$. As a second step, the mixture of the two latent codes $X'$, is also passed through the decoder $g$, which reconstructs the input instances by decoding the representation $X'$. The output from the decoder can be expressed as:

\begin{equation}
    (\hat{X}, p) = (g(X'; \theta_g), p)
\end{equation}

where $\theta_g$ refers to the model parameters. 
Unlike traditional decoders, $g$ is also given access to $p$. This allows it to provide an approximate reconstruction $(\hat{X}, p) = (g(X'; \theta_g), p)$ of the input $X'$ and the original $p$. This ensures a lower reconstruction error even after the representations have been disentangled. 

In order to generate semantically meaningful interpolations, our model additionally takes the help of an interpolation critic $Critic_i$ in its third step. This critic tries to recover the interpolation co-efficient $\alpha$, from the output of the decoder $(\hat{X}, p)$. During training, the network tries to constantly fool the critic to output $\alpha=0$, implying that $(\hat{X}, p)$ is non-interpolated. The inclusion of such a loss has been beneficial in learning the data manifold~\cite{berthelot2018understanding}. The loss that the interpolation critic minimizes is given by: 
\begin{equation}
    L_{Critic_{i}} = ||\alpha-\hat{\alpha}||_{2}^2
    \label{eq:critic_i}
\end{equation}

The autoencoder's overall loss function is then modified by adding the following regularization terms: 
\begin{equation}
\begin{split}
     L_{ae} = & \underbrace{||X- g(X')||_{2}^2}_\text{Reconstruction Error} + \underbrace{\beta ||p- Critic_{dis}(X')||_{2}^2}_\text{Protected Disentanglement} \\+ & \underbrace{\lambda||\alpha-Critic_i(g(X'), p)||_{2}^2}_\text{Interpolation Error}
    \label{eq:ae}
\end{split}
\end{equation}

where $\beta$ is the weight given to determine the importance of disentanglement with respect to the reconstruction, and $\lambda$ is the weight of the regularization term for determining the importance of interpolation. Algorithm \ref{algo} presents the overall methodology followed by FRIED to learn the latent features $X'$ and the reconstruction $\hat{X}$.

\begin{algorithm}
\SetAlgoLined
\SetNoFillComment
\textbf{Input:} Features $X$; Protected Attribute(s) $p$;\\
\textbf{Output:} Disentangled Representation $X'$; Mixed Reconstruction $\hat{X}$\\
\SetAlgoLined
//~~\CommentSty{Training}\\
\While{not converged}{
$\alpha \xleftarrow{} rand(0, 1)$ \\
$X' \xleftarrow{} Interpolate(\alpha, f(X_k, p), f(X_{k+1}, p)) \forall X_k \in X$\\
$\hat{p} \xleftarrow{} Critic_{dis}(X')$\\ 
$(\hat{X}, p) \xleftarrow{} (g(X'), p)$\\
$\hat{\alpha} \xleftarrow{} Critic_i(\hat{X}, p)$\\
Compute loss using Eqs. (\ref{eq:critic_dis}), (\ref{eq:critic_i}), and (\ref{eq:ae})\;
}

//~~\CommentSty{Testing}\\
$X' \xleftarrow{} f(X, p)$\\
$\hat{X} \xleftarrow{} g(X', p)$\\
\Return{$X', \hat{X}$}
\caption{FRIED: Fair Representation Learning using Interpolation Enabled Disentanglement}
\label{algo}
\end{algorithm}

\subsection{Model Auditing}
\begin{figure}[h]
    \includegraphics[width=0.48\textwidth]{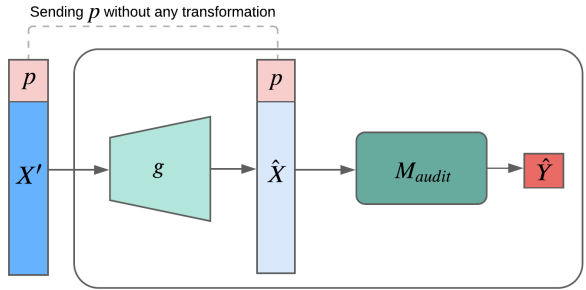}
    \caption{Architecture for model auditing: Indirect influence of $p$ on $M_{audit}$ is same as the direct influence of $p$ on the decoder $g$ and $M_{audit}$ combined.}
    \label{fig:model_audit1}
\end{figure}

Our setup for model auditing presented in Fig. \ref{fig:model_audit1} slightly differs from that in Fig~\ref{fig:architecture}. We recall Proposition 1 from Marx et. al. \cite{marx2019disentangling}, which essentially states that the indirect influence of the feature $p$, on the model $M_{audit}$, is same as the direct influence of $p$ on the decoder $g$, and the target model $M_{audit}$ combined (Fig. \ref{fig:model_audit1}). Although $p$ is not given to the decoder $g$ during training, it is passed through the decoder without any transformation during model auditing.

\section{Improving Fairness-Accuracy Trade-off}

\label{sec:friedtheory}
In this section, for the sake of brevity, we confine ourselves to the case of binary classification problems for the theory developed henceforth. Without loss of generality, let $p=0$ be the unprotected group and $p=1$ be the protected group. Let the features in the given dataset have the following distributions: $X|_{Y=0,p=0} {\sim} P_0(x)$ and $X|_{Y=1,p=0} {\sim} P_1(x).$ Similarly, $X|_{Y=0,p=1} {\sim} Q_0(x)$ and $X|_{Y=1,p=1} {\sim} Q_1(x).$ We use the notation $H_0$ and $H_1$ to denote the hypothesis that the true label is $0$ or $1$.

We now re-state a recently proposed result in Dutta et. al.~\cite{sanghamittrachernoff} which helps us in theoretically proving and empirically quantifying how the latent features learned from FRIED improves the fairness-accuracy trade-off. The theorem mainly demonstrates how gathering more features (in our case $X'$ learned from FRIED) can help in improving the Chernoff information (defined below) of the protected group without affecting that of the unprotected group. Gathering more features helps classify members of the protected group more carefully with additional separability information that was not present in the initial dataset ($X$).

Let $X'$ denote the additional features so that now $(X,X')$ is used for classification of the unprivileged group. Let the features $(X,X')$ have the following distributions: $(X,X')|_{Y=0,p=0}\sim W_0(x,x')$ and $(X,X')|_{Y=1,p=0}\sim W_1(x,x')$. Note that, $P_0(x)=\sum_{x'}W_0(x,x')$ and $P_1(x)=\sum_{x'}W_1(x,x')$. 

The separability of the positive and negative labels of a group in the given dataset is quantified using a tool from information theory called \emph{Chernoff information}.
\begin{lem}
\label{lem:separability}
Given two hypotheses, $H_0:P_0(x)$ and $H_1:P_1(x)$, the Chernoff exponent of the Bayes
optimal classifier is given by the Chernoff information: \begin{align}\mathrm{C}(P_0,P_1)=- \min_{ u \in (0,1)} \log{\left(\sum_{x} P_0(x)^{1-u}P_1(x)^{u}\right)}. \nonumber \end{align}
\end{lem}

The primary goal here is to re-state the conditions under which the separability improves with addition of more features, i.e., $\mathrm{C}(W_0,W_1)>\mathrm{C}(P_0,P_1)$. 

\begin{thm}[Improving Separability]
The Chernoff information $\mathrm{C}(W_0,W_1)$ is strictly greater than $\mathrm{C}(P_0,P_1)$ if and only if $X'$ and $Y$ are not independent of each other given $X$ and $p=0$, i.e., the conditional mutual information (CMI) $I(X';Y|X,Z=0)>0$.
\label{thm:explainability}
\end{thm}

We suggest readers to refer to the SM for the proof of this based on Theorem 3 from here~\cite{sanghamittrachernoff}.  We now need an empirical estimate for CMI which is a non-trivial problem. Estimating ($I(X;Y|Z)$) from finite samples can be stated as follows. Let us consider three random variables $X$, $Y$, $Z \sim p(x,y,z)$, where $p(x,y,z)$ is the joint distribution (not to be confused with protected attribute $p$). Let the dimensions of the random variables be $d_x$, $d_y$ and $d_z$ respectively. We are given $n$ samples $\{(x_i, y_i, z_i) \}_{i=1}^n$ drawn i.i.d. from $p(x,y,z)$. So $x_i \in \mathbb{R}^{d_x}, y_i \in \mathbb{R}^{d_y}$ and $z_i \in \mathbb{R}^{d_z}$. The goal is to estimate $I(X;Y|Z)$ from these $n$ samples. We now look at the connection between CMI and KL divergence.

\subsection{Divergence Based CMI Estimation}
\begin{defn}
The Kullback-Leibler (KL) divergence between two distributions $p(\cdot)$ and $q(\cdot)$ is given as :
\[
D_{KL}(p||q) = \int p(x) \log \frac{p(x)}{q(x)} \, dx
\]
\end{defn}

\begin{defn}
Conditional Mutual Information (CMI) can be expressed as a KL-divergence between two distributions $p(x,y,z)$ and $q(x,y,z) = p(x,z)p(y|z)$, i.e., 
\[
I(X;Y|Z) = D_{KL}(p(x,y,z)||p(x,z)p(y|z))
\]
\end{defn}

\begin{defn}
The Donsker-Varadhan representation expresses KL-divergence as a supremum over functions,
\begin{equation}
D_{KL}(p||q) = \sup\limits_{f \in \mathcal{F}} \mathop{\mathbb{E}}\limits_{x\sim p}[ f(x)] - \log(\mathop{\mathbb{E}}\limits_{x\sim q} [\exp(f(x))])
\label{DV-bound}
\end{equation}
\end{defn}

where the function class $\mathcal{F}$ includes those functions that lead to finite values of the expectations. 

\subsection{Difference Based CMI Estimation}
An intuitive approach to estimate CMI could be to express it as a difference of two mutual information terms as presented in~\cite{mukherjee20a} by invoking the chain rule, i.e.: $I(X;Y|Z) = I(X; Y,Z) - I(X;Z)$.
Since mutual information is a special case of KL-divergence, viz. $I(X;Y) = D_{KL}(p(x,y)||p(x)p(y))$, this again calls for a empirical KL-divergence estimator which is presented below.

Given $n$ i.i.d. samples $\{x_i^p\}_{i=1}^n , x_i^p \sim p(x)$ and $m$ i.i.d. samples $\{x_j^q\}_{j=1}^m, x_j^q \sim q(x)$, we want to estimate $D_{KL}(p||q)$. We label the points drawn from $p(\cdot)$ as $y = 1$ and those from $q(\cdot)$ as $y=0$. A binary classifier is then trained on this supervised classification task. Let the prediction for a point $l$ by the classifier is $\gamma_l$ where $\gamma_l = Pr(y = 1 | x_l)$ ($Pr$ denotes probability). Then the point-wise likelihood ratio for data point $l$ is given by $\mathcal{L}(x_l) = \frac{\gamma_l}{1 -\gamma_l}$. 

The following Proposition is elementary and has already been observed here~\cite{belghazi2018mine}. We re-state it here for completeness and quick reference. 

\begin{prop} The optimal function in Donsker-Varadhan representation \eqref{DV-bound} is the one that computes the point-wise log-likelihood ratio, i.e, $f^*(x) =  \log \frac{p(x)}{q(x)} \, \forall \, x$, (assuming $p(x) = 0$, where-ever $q(x) = 0$).
\label{prop-1}
\end{prop}
Plugging in the definition of $f^{*}(x)$ in the Donsker-Varadhan representation \eqref{DV-bound}, we obtain 
\begin{align*}
&\mathop{\mathbb{E}}\limits_{x\sim p}\left[ \log \frac{p(x)}{q(x)} \right] - \log \left(\mathop{\mathbb{E}}\limits_{x\sim q} [\exp(\log \frac{p(x)}{q(x)})] \right) \\
&= D_{KL}(p||q) - \log \left( \sum\limits_{x \in \mathcal{X}} q(x) \frac{p(x)}{q(x)}\right) = D_{KL}(p||q)
\end{align*}

Based on Proposition \ref{prop-1}, the next step is to substitute the estimates of point-wise likelihood ratio in \eqref{DV-bound} to obtain an estimate of KL-divergence.
\begin{equation}
\hat{D}_{KL}(p||q) = \frac{1}{n}\sum\limits_{i=1}^n \log \mathcal{L}(x_i^p) - \log \left(  \frac{1}{m}\sum\limits_{j=1}^m \mathcal{L}(x_j^q) \right) \nonumber
\label{plug-in}
\end{equation}

With the help of this KL-divergence estimator the difference-based algorithm~\cite{mukherjee20a} can be used for computing the CMI. Empirical CMI values obtained for FRIED using this algorithm are provided in the SM.

\section{Experimental Results}\label{exp}

\begin{figure*}[h]
    \begin{subfigure}[]{0.33\linewidth}
    \includegraphics[width=\textwidth]{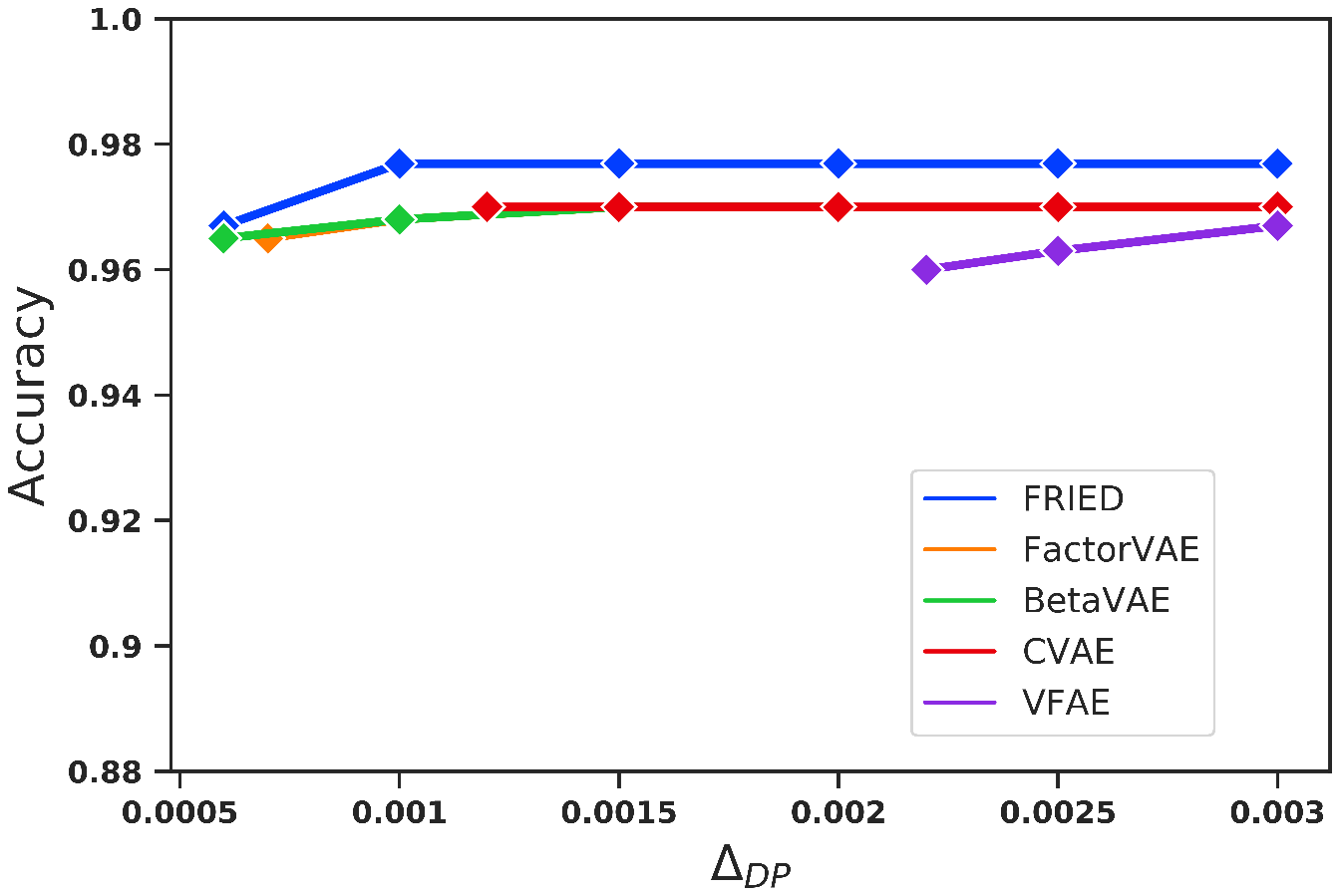}
    \caption{Protected Attribute: Scale}
    \end{subfigure}
    \begin{subfigure}[]{0.33\linewidth}
    \includegraphics[width=\textwidth]{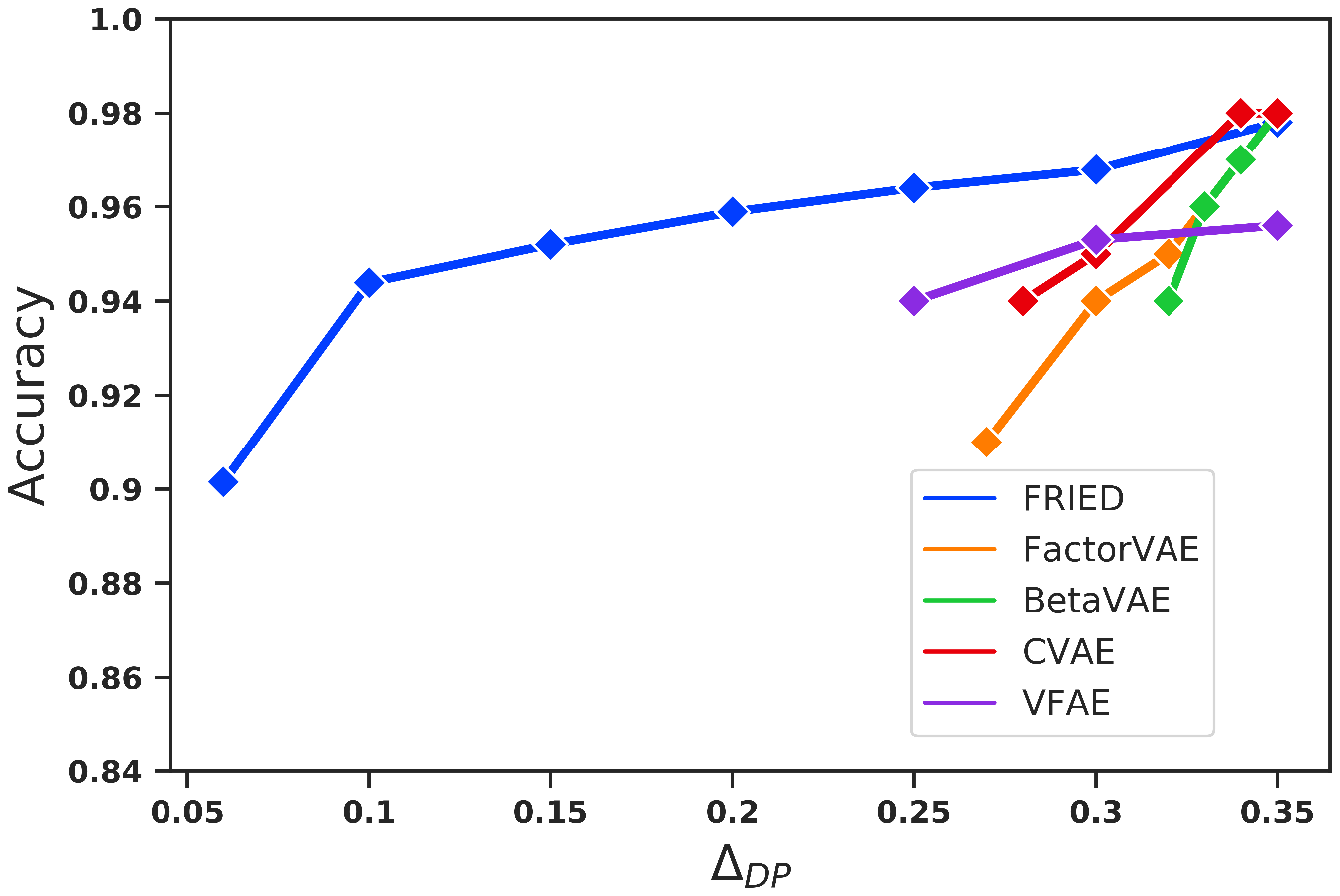}
    \caption{Protected Attribute: Shape}
    \end{subfigure}
    \begin{subfigure}[]{0.33\linewidth}
    \includegraphics[width=\textwidth]{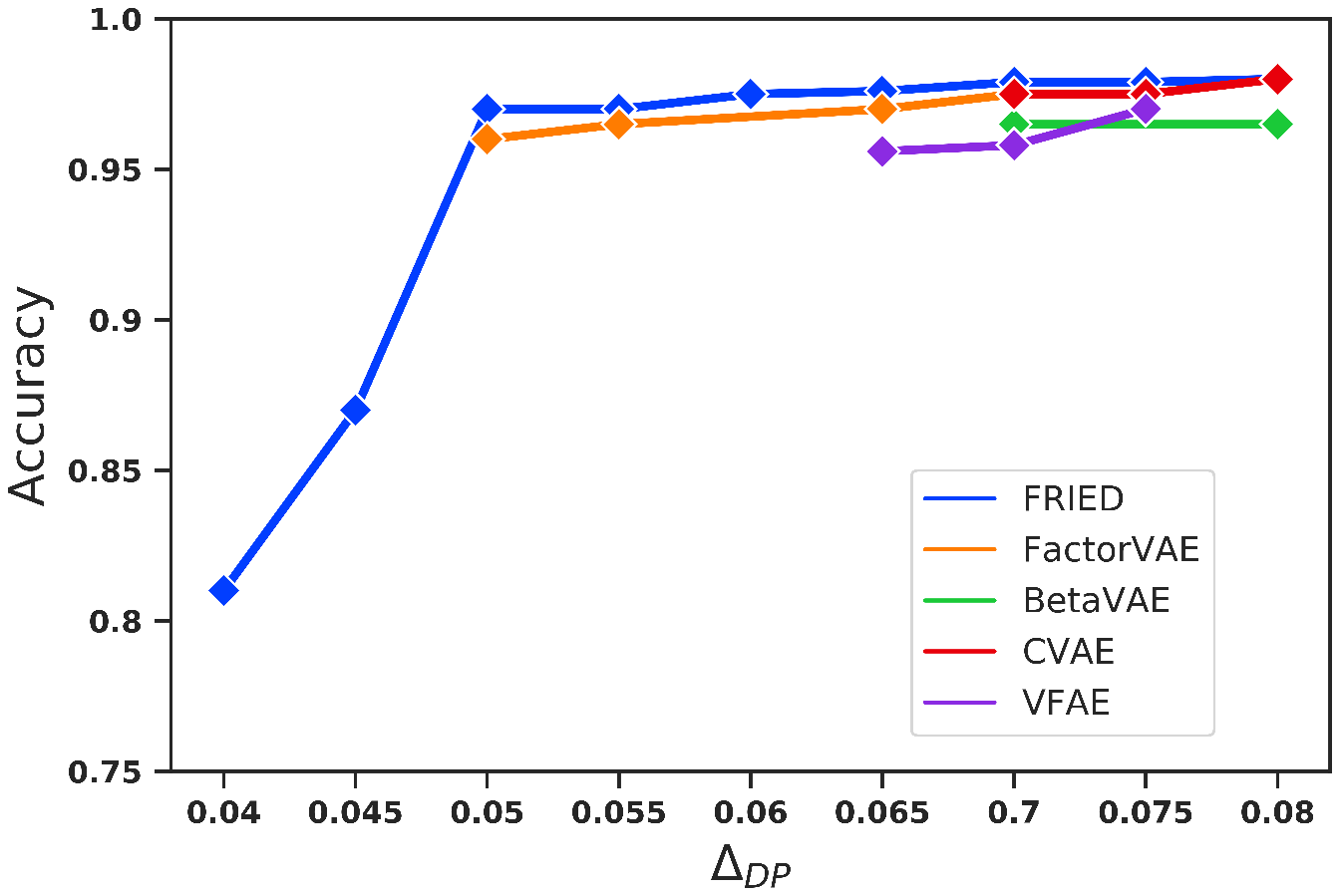}
    \caption{Protected Attribute: Shape $\land$ Scale}
    \end{subfigure}
    \begin{subfigure}[]{0.33\linewidth}
    \includegraphics[width=\textwidth]{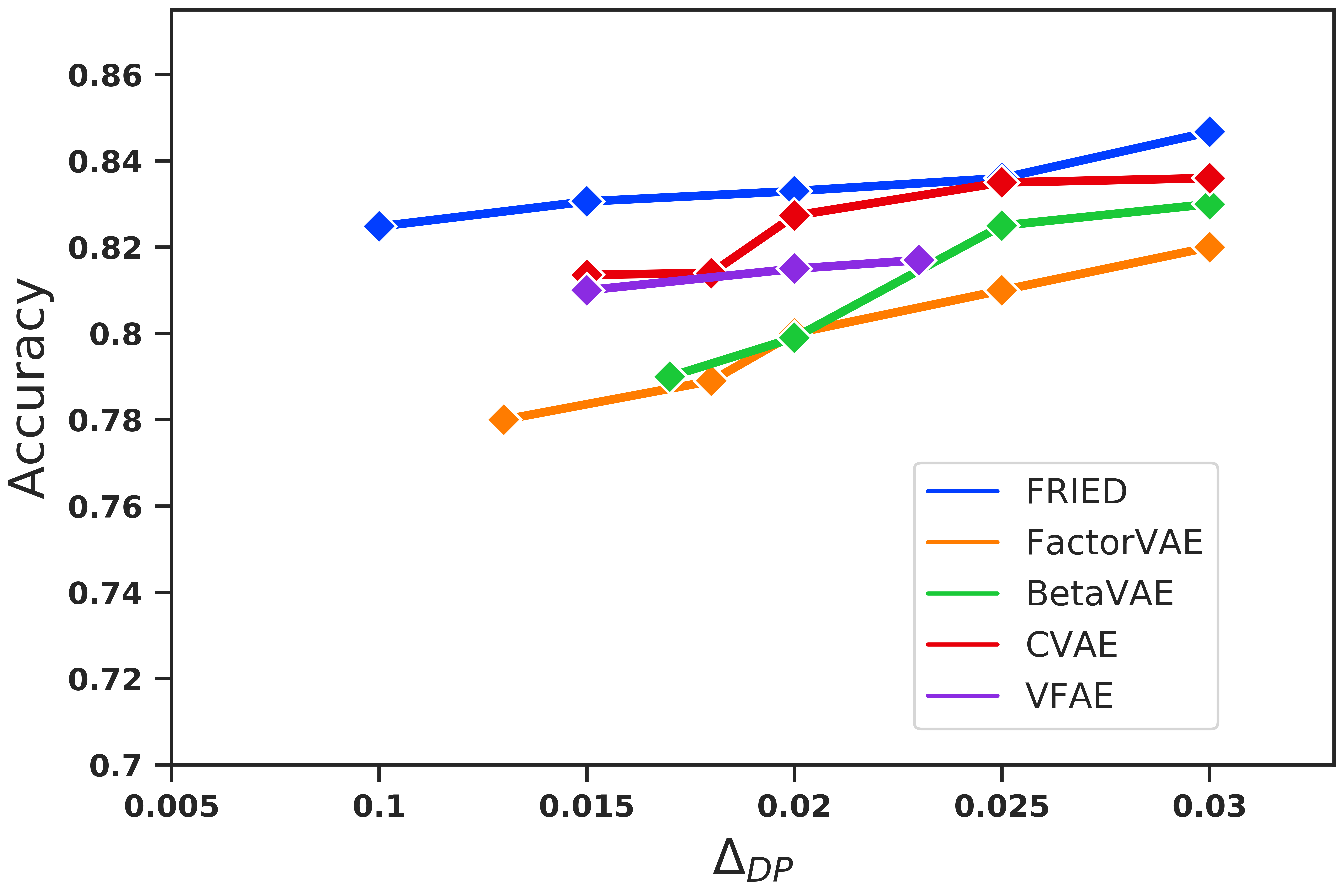}
    \caption{Protected Attribute: Gender}
    \end{subfigure}
    \begin{subfigure}[]{0.33\linewidth}
    \includegraphics[width=\textwidth]{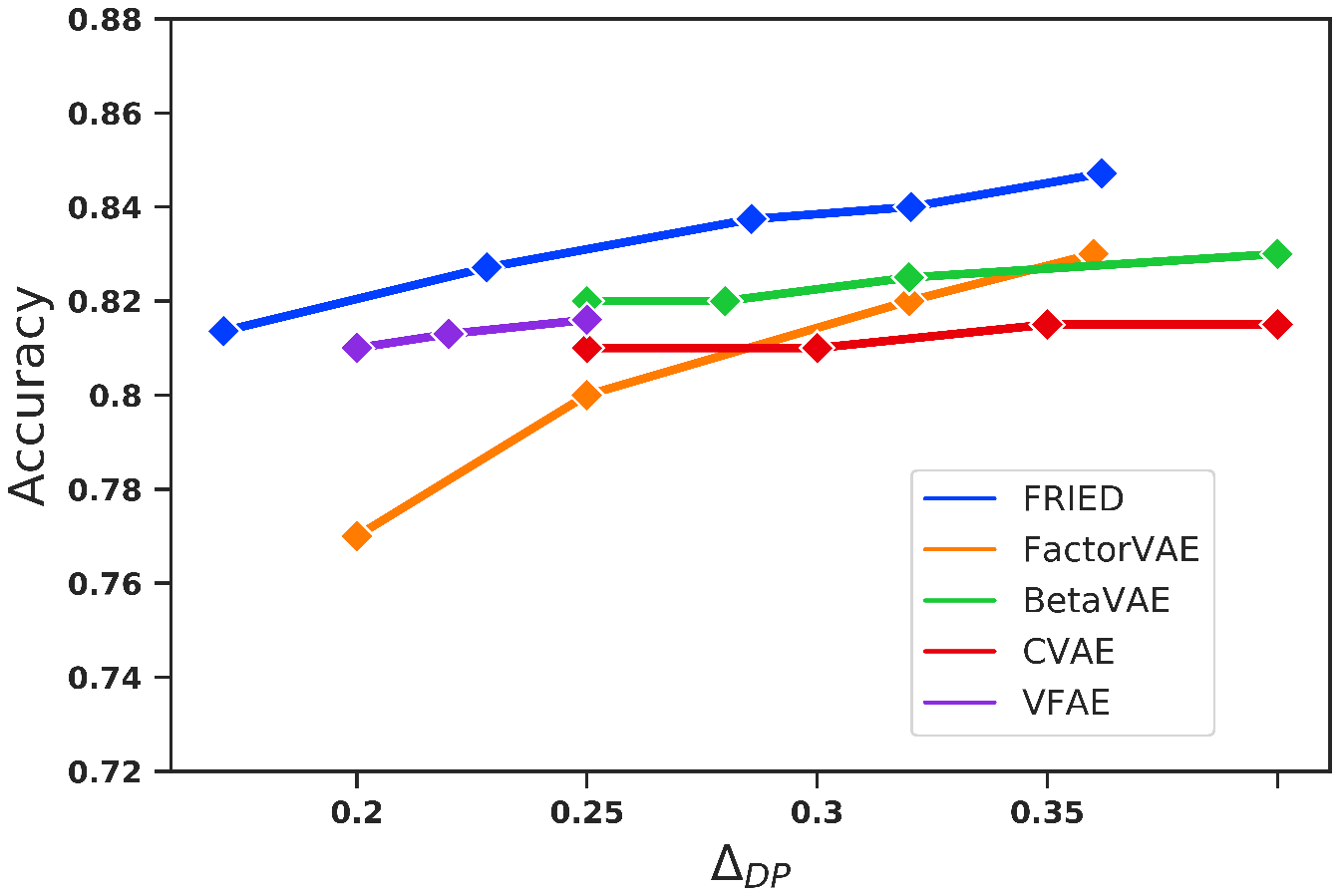}
    \caption{Protected Attribute: Race}
    \end{subfigure}
    \begin{subfigure}[]{0.33\linewidth}
    \includegraphics[width=\textwidth]{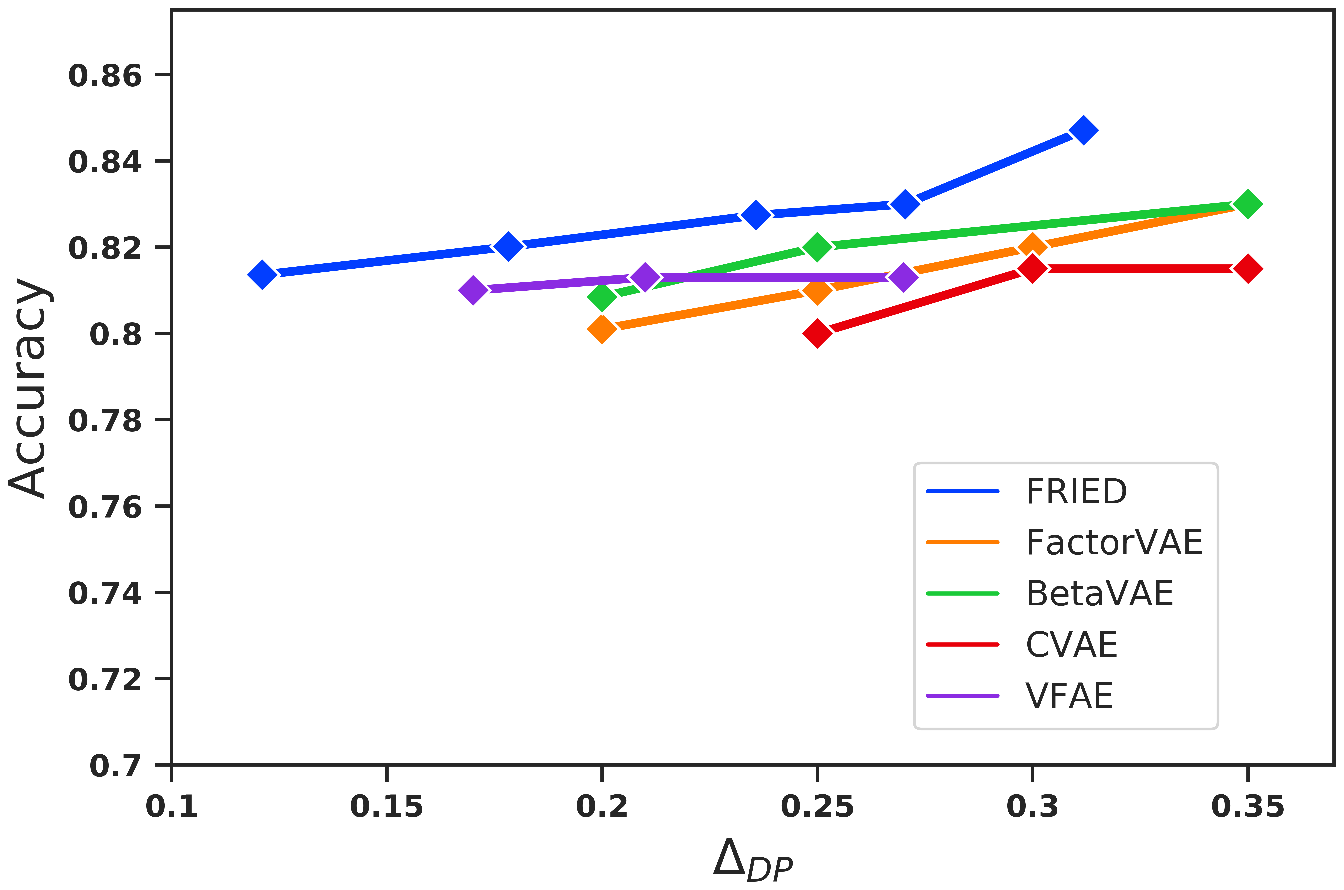}
    \caption{Protected Attribute: Gender $\land$ Race}
    \end{subfigure}
    \caption{Fairness Accuracy Trade-Off curves for dSprites (top) and Adult Dataset (bottom) for 3 different protected attributes. We compare FRIED's fairness and accuracy with four different models (i) FactorVAE, (ii) BetaVAE,  (iii) Conditional VAE (CVAE), and (iv) Variational Fair Autoencoder (VFAE). We sweep a range of hyperparameters to generate the Pareto fronts. The X-axis represents the Demographic Parity Difference ($\Delta_{DP}$) between different groups. Lower the value of $\Delta_{DP}$, fairer the model. Y-axis represents the downstream classification accuracy. Top right is best accuracy and lowest fairness. Bottom left represents best fairness and lowest accuracy. FRIED outperforms the other model obtaining the best fairness-accuracy trade-off while being robust to changes in hyperparameters.}
    \label{fig:data_fair}
\end{figure*}

In this section, we evaluate the fairness of the learned representation $X'$ along with $X$ and measure downstream classification accuracy for FRIED and several other baselines. We also conduct an ablation study to evaluate importance of different components within FRIED.

\subsection{Dataset Description}
We conduct experiments on the following real-world datasets to demonstrate the efficacy of the proposed FRIED model: 
\begin{itemize}[leftmargin=*, nosep]
    \item {\textbf{UCI Adult \cite{Dua:2019}}}: It contains census information of ~40,000 people including details such as age, work class, education, sex, marital status, race, native country, the number of hours worked, etc. The task is to predict whether the income $\geq$ 50k.
    \item {\textbf{COMPAS Recidivism \cite{angwin2016machine}}}:
    COMPAS is a commercial algorithm used by the judges, probation and parole officers across the United States to assess a defendant's likelihood to re-offend. The task is to predict a defendant's likelihood of violent recidivism in 2 years.
    \item{\textbf{dSprites Datatset \cite{dsprites17}:}}
    It is a dataset of 2D shapes generated from 6 ground truth latent factors: color, shape, scale, rotation, x and y positions of a sprite. It contains 737,280 total images. The task involves predicting the Y position of a sprite.
    \item{\textbf{Wikipedia Toxic Comments \cite{Thain2017}:}}
    This is a dataset of 100,000 English Wikipedia comments which have been labelled by human annotators according to varying levels of toxicity. The aim is to predict the toxicity of the comments.
\end{itemize}

Additionally, along with an expert, we audit an expert provided black-box classification model on the following Opioid healthcare claims dataset to generate meaningful insights.
\begin{itemize}[leftmargin=*, nosep]
 \item {\textbf{Opioid \citep{zhang2017exploring}:}}
    The Opioid dataset used for our experiments is a sample from the Truven Health Analytics MarketScan{\textregistered} Commercial database that contains information of 80 million enrollees. More information about the dataset can be found in  \citep{zhang2017exploring}.  The classification task is to predict long-term opioid addiction of a patient.
\end{itemize}

\noindent\textbf{Implementation Details} The critical hyper-parameters such as the number of nodes in the hidden layer, dimensions of the latent vector, learning rate, and other parameters were varied according to the dataset. For the UCI Adult dataset, each of the above components consisted of two hidden layers of 30 and 15 hidden units. The latent vector had 30 dimensions. The model was trained for 100 epochs with a mini-batch of size 100, using stochastic gradient descent (SGD) with a learning rate of 0.01. The COMPAS model was trained similar to the UCI Adult model. Two hidden layers of size 25 and 12 and a latent vector of size 12 were used for training the model for 75 epochs. For the dSprites dataset, only images with scale greater than 0.7 were considered. The images were flattened out and fed to the model with two hidden layers of size 256 and 64. The size of the latent vector was 32. The model was trained for 5,000 epochs with a batch-size of 50 and a learning rate of 0.03. The model for Wikipedia Toxic Comments data used bag of words as its input features. The data was pre-processed and the size of the vocabulary was limited to 10,000 words. It was trained using a batch size of 100 for 7,000 epochs with a learning rate of 0.05 and three hidden layers of size 500, 250 and 100; the size of the latent vector was 50. The classification model for all the above datasets was a two-layer multi-layer perceptron. The train-test split was 80-20 for all datasets and the results reported are for 5 folds.

\subsection{Fairness-Accuracy Trade Off}
To evaluate the fairness of the model, we use \textit{Demographic Parity Difference} ($\Delta_{DP}$) \cite{calders2010three}.
Demographic Parity states that the prediction should be similar across different groups irrespective of the sensitive attribute. We define $\Delta_{DP}$ as the absolute difference in the demographic parity between two groups. It is given by:
    \begin{equation}
      \Delta_{DP} = |{P(\hat{Y}|p=0)}-{P(\hat{Y}|p=1)}|
    \end{equation}

We compare FRIED's performance with several baselines, namely, (i) FactorVAE \cite{kim2018disentangling}, (ii)  BetaVAE \cite{higgins2016beta}, (iii) CVAE \cite{sohn2015learning}, and (iv) Variational Fair AE (VFAE) \cite{louizos2015variational}. The results for the UCI Adult and the dSprites datasets are shown in Fig. \ref{fig:data_fair}. The results were obtained after sweeping the values of the hyperparameters $\beta$ and $\lambda$ for $Critic_{dis}$ and $Critic_{i}$ respectively (shown in Eq.~\ref{eq:ae}). The values were varied from 0 to 1 and the corresponding accuracy and the $\Delta_{DP}$ of the model were plotted (as seen in Fig. \ref{fig:data_fair}). These results also show that the model is robust to changes in hyperparameters. We observe that FRIED achieves better fairness while preserving accuracy, thereby, achieving improved fairness-accuracy trade-off compared to other baselines.

Our intuition here for the better performance of FRIED in terms of the improved fairness-accuracy trade-off is supported through the theory presented in Section~\ref{sec:friedtheory}. We compute the empirical CMI values using the difference-based algorithm~\cite{mukherjee20a} with a MLP classifier (architecture and parameter details in SM). The computed CMI value ($I(X';Y|X,Z=0)$) for all datasets with the latent code generated from FRIED was found to be positive and approximately 20\% higher compared to other methods. As CMI values are greater than 0 (exact values in SM), from Theorem~\ref{lem:separability}, we can conclude that the representation learned from FRIED helps in improving the fairness-accuracy trade-off significantly compared to other baselines.

\subsection{Model Auditing Case Study}\label{audit}

\begin{figure}[h!]
    \begin{subfigure}[]{\linewidth}
    \includegraphics[width=\textwidth]{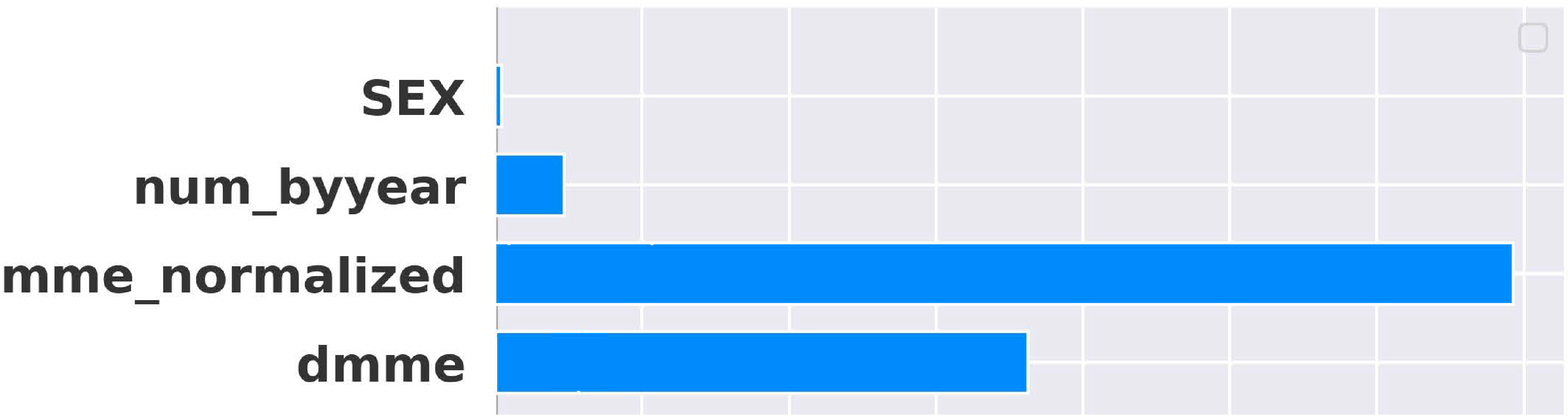}\\
    \includegraphics[width=\textwidth]{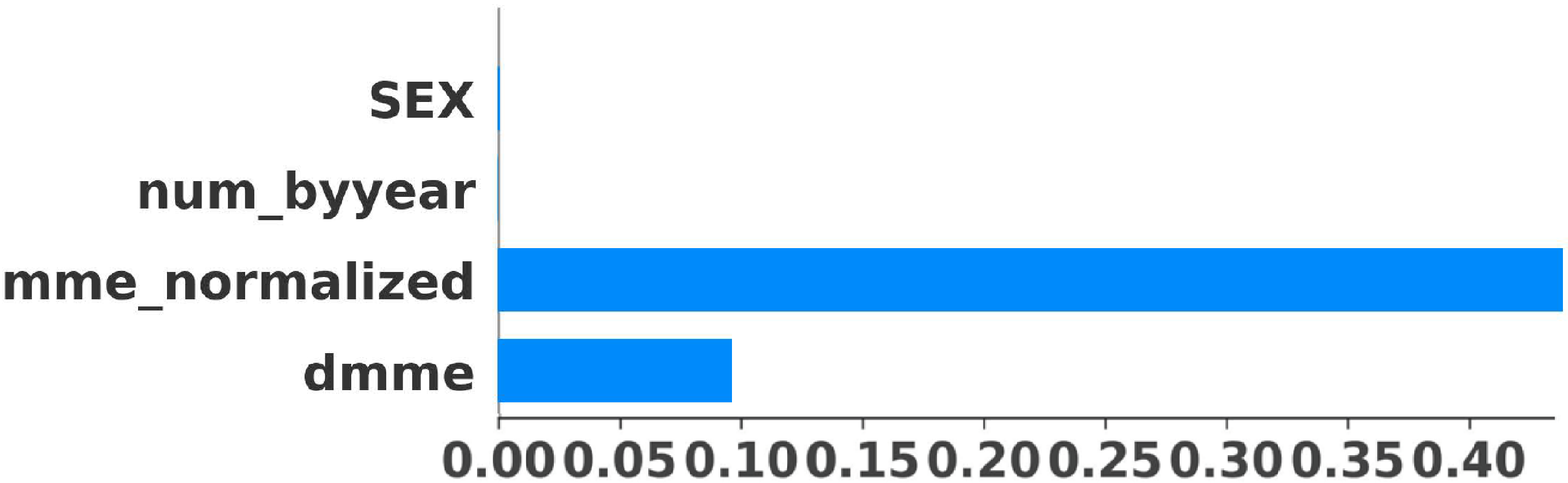}
    \caption{FRIED (tiled background) inferred that daily morphine milligram equivalents (dmme) and healthcare utilization per year (num\_byyear) had a high indirect influence on opioid addiction. The direct influence plot (plain background) shows that the original representation could not highlight the important factors of `dmme' and `num\_byyear'.}
    \label{fig:indirect_opioid1}
    \end{subfigure}
    \begin{subfigure}[]{\linewidth}
    \includegraphics[width=\textwidth]{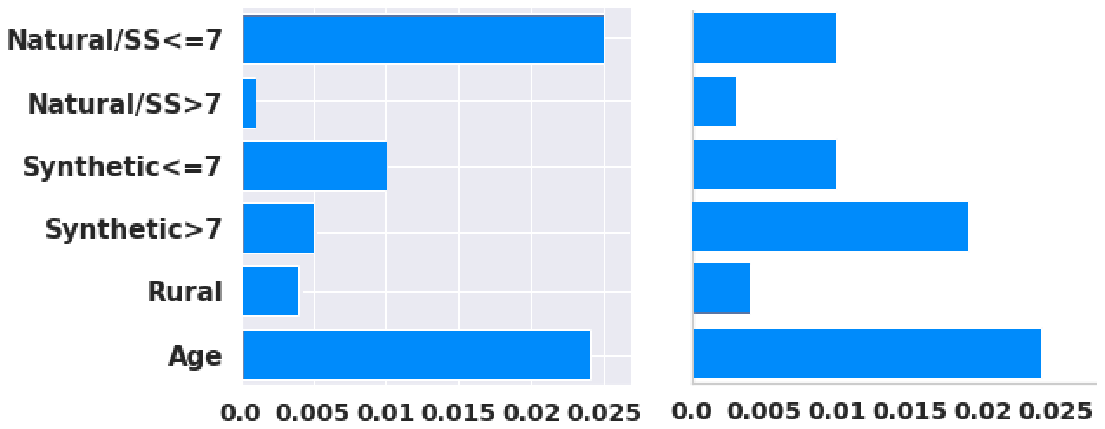}
    \caption{FRIED (tiled background) inferred that natural or semi-synthetic (SS) opioids $\leq$ 7 days had a high indirect influence on opioid addiction. The direct influence plot (plain background) captured the misleading global addiction pattern of synthetic opioids instead.}
    \label{fig:indirect_opioid2}
    \end{subfigure}
    \caption{Auditing an opioid addiction risk prediction classifier. SHAP plots comparing the indirect feature importance w.r.t. \textit{Rural} and \textit{Age} as protected (sensitive) attributes inferred using our model FRIED (tiled background) vs. the direct influence plot (plain background) from the original representations on the Opioid Healthcare Dataset.}
\end{figure}

In this experiment, we demonstrate our model's ability to audit a black-box model for identifying proxy features. We evaluate the indirect influence of the protected attributes by auditing the model $M_{audit}$ using FRIED to obtain $SHAP$ values \cite{lundberg2017unified} based insights. $SHAP$ values are game-theoretic explanations that help identify the feature influence by averaging the marginal contribution of each feature over all possible input sequences and their corresponding predictions. 

\begin{table*}[h]
    \caption{Ablation study reporting the Informativeness Score with standard deviation computed over 5 folds for our model (FRIED) compared with (i) Vanilla Autoencoder \cite{kramer1991nonlinear}, (ii) FRIED w/o $Critic_{dis}$, and (iii) FRIED w/o $Critic_i$. The values in bold represent best performance. This demonstrates that interpolation aids in disentanglement of protected attributes in latent space.}
    \label{tab:dis}
    \centering
    \begin{tabular}{|l|c|c|c|c|}
        \hline
        \textbf{Dataset} &  \textbf{Autoencoder} & \textbf{FRIED w/o $Critic_{dis}$} & \textbf{FRIED w/o $Critic_{i}$} & \textbf{FRIED}\\
        \hline
        Adult & 0.9466 $\pm$ 0.002 & 0.9511 $\pm$ 0.003 & 0.9401 $\pm$ 0.001 & \textbf{0.9661 $\pm$ 0.001}  \\
        \hline
        COMPAS & 0.7741 $\pm$ 0.007 & 0.7227 $\pm$ 0.005 & 0.7790 $\pm$ 0.006 & \textbf{0.8261 $\pm$ 0.008} \\
        \hline
        dSprites & 0.6901 $\pm$ 0.053 & 0.7307 $\pm$ 0.005 & 0.6705 $\pm$ 0.014 & \textbf{0.7744 $\pm$ 0.001}\\
        \hline
        Wikipedia & 0.9720 $\pm$ 0.001 & 0.9719 $\pm$ 0.003 & 0.9724 $\pm$ 0.002 & \textbf{0.9730 $\pm$ 0.001}\\
        \hline
    \end{tabular}
\end{table*}

\subsubsection{\textbf{Auditing a black-box classifier which predicts risk of opioid addiction}}
\label{sec:expertstudy}

To demonstrate FRIED's impact, we conducted a model auditing case study involving a real-world opioid healthcare claims dataset with the aim of predicting short or long-term opioid addiction. The black-box classifier that was being audited was provided to us by the expert. Expert informed us that old patients from rural areas are over-prescribed opioids compared to their urban counterparts. Long drive times from rural areas to opioid rehabilitation centers have mainly exacerbated this epidemic with rural older adults being most affected~\cite{edelman2018rural}.  They had higher levels of daily morphine milligram equivalents (dmme) and higher healthcare visits (num\_byyear) than usual. The direct features in this case are \textit{`Rural'} and \textit{`Age}' whereas the indirect features are `dmme' and `num\_byyear' (healthcare visits). The ideal goal while auditing here should be to uncover the influence of these indirect features on the model outcome while \textit{disentangling} the effect of the direct features.

Auditing the original representations, as shown in Fig~\ref{fig:indirect_opioid1}, was unable to highlight the influence of the `dmme' and `num\_byyear' (healthcare visits). Another major insight the expert derived was that after applying FRIED, the influence of whether a patient has taken natural or semi-synthetic opioids for $\le$ 7 days was more evident on the model. This was not picked up clearly with the original representation which highlighted synthetic opioids as shown in Fig~\ref{fig:indirect_opioid2}. A study \footnote{https://www.cdc.gov/nchs/products/databriefs/db345.htm} shared by the expert corroborated these results that addiction rates were higher in rural counties than in urban counties for drug overdose deaths involving natural and semi-synthetic opioids (4.9 vs 4.3). Synthetic opioids such as Fentanyl are known to have exacerbated the opioid epidemic across the United States; however, FRIED was able to tease out the importance of short-term natural and semi-synthetic opioids compared to synthetic opioids in rural areas which was what convinced the expert on our method. The SHAP plots for all other datasets can be found in the supplementary material.

\subsection{Ablation Study}\label{sec:trivial_baseline}

We conduct an ablation study with two protected attributes to quantitatively demonstrate the importance of various components on four different datasets. Race and gender were treated as protected attribute for the Adult and COMPAS datasets, shape and scale were protected attributes for dSprites, and sexuality and race were the protected attributes for the experiments on the Wikipedia dataset. The various components were as follows:

\begin{itemize}[leftmargin=*, nosep]
    \item \textbf{Vanilla Autoencoder (AE) \cite{kramer1991nonlinear}}: A vanilla autoencoder that attempts to reconstruct the input without using any critic.  
    \item  \textbf{FRIED w/o $Critic_{dis}$}: Uses interpolation critic for reconstructing the input but does not employ the disentanglement critic.
    \item \textbf{FRIED w/o $Critic_i$}: Uses a disentanglement critic but does not use an interpolating critic for reconstruction.
\end{itemize}

Disentangled representations have been shown to be fairer \cite{locatello2019fairness}. Therefore, better disentanglement implies better fairness. We conduct experiments to evaluate the quality of the learned disentangled representations using the
\textbf{Informativeness Score} \cite{do2019theory}. \textit{Informativeness} of the representations is quantified by the mutual information between the a particular representation $z_i$ and data $x$. For a representation $z$ to be useful it should be representative of the data $x$. It is given by: $I(x, z_i) = \int_x \int_z p_D(x)q(z_i|x)log\frac{q(z_i|x)}{q(z_i)}dz~dx$ where $q(z_i) = \int_x p_D(x)q(z_i|x) dx$. Larger the value, better the informativeness. The informativeness value is also a measure of the model's ability to appropriately disentangle the underlying factors of variation. The results for the ablation study with (i) a vanilla autoencoder (AE), (ii) FRIED w/o $Critic_{dis}$, and (iii) FRIED w/o $Critic_{i}$ are shown in Table \ref{tab:dis}. We observe that FRIED has better informativeness because of learning the manifold through interpolation implying improved disentanglement, which in turn results in better fairness. FRIED demonstrates a much higher informativeness score of \textbf{0.7744} even on the large dSprites image dataset consisting of 737,280 images.

\section{Conclusion}\label{concl}

In this paper, we presented FRIED, a novel approach to learn fair representations using interpolation enabled disentanglement. FRIED made use of an autoencoder along with adversarially constrained interpolation to semantically mix features in the latent space. The representations learnt are fairer also help in improving downstream classifier performance. We present novel theory on how we improve the fairness-accuracy trade-off using mutual information theory.
Additionally, we conducted an expert study to audit an opioid addiction risk prediction classifier on a healthcare claims dataset. The expert concluded that the representation learned from FRIED made it conducive for the model to tease out indirect influence patterns from the cohort.  For future work, we plan to better understand the effects of the features learned through latent interpolation on improving the fairness-accuracy trade-off. We also plan to experiment with more than two points for interpolation.


\bibliographystyle{abbrv}
{\footnotesize
\bibliography{sample-base}}



\end{document}